\documentclass[conference]{IEEEtran}
\IEEEoverridecommandlockouts
\usepackage{cite}
\usepackage{amsmath,amssymb,amsfonts}
\usepackage{algorithmic}
\usepackage{graphicx}
\usepackage{mwe}      
\usepackage{textcomp}
\usepackage{xcolor}
\usepackage{csquotes}
\usepackage{url}
\usepackage{tikz}
\usetikzlibrary{shapes, arrows}
\usetikzlibrary{trees}
\usepackage{booktabs}
\usepackage{multirow}
\usetikzlibrary{positioning, fit, shapes.geometric}
\usepackage{caption}
\usepackage{subcaption}
\usepackage{amsmath}
\usepackage{mathtools}
\usepackage{pgfplots}
\pgfplotsset{compat=newest}
\usepackage{hyperref}
\usepackage{makecell}
\usepackage{array}
\usepackage{xcolor}
\usepackage{adjustbox}
\usepackage{siunitx}
\usepackage{geometry}

\newcommand\copyrighttext{%
	\footnotesize \copyright 2024 IEEE. Personal use of this material is permitted. Permission from IEEE must be obtained for all other uses, in any current or future media, including reprinting/republishing this material for advertising or promotional purposes, creating new collective works, for resale or redistribution to servers or lists, or reuse of any copyrighted component of this work in other works.}
\newcommand\copyrightnotice{%
	\begin{tikzpicture}[remember picture,overlay]
		\node[anchor=south,yshift=10pt] at (current page.south) {\fbox{\parbox{\dimexpr\textwidth-\fboxsep-\fboxrule\relax}{\copyrighttext}}};
	\end{tikzpicture}%
}

\definecolor{LightBlue}{RGB}{173,216,230}

\def\BibTeX{{\rm B\kern-.05em{\sc i\kern-.025em b}\kern-.08em
    T\kern-.1667em\lower.7ex\hbox{E}\kern-.125emX}}

\begin{document}
\renewcommand{\IEEEtitletopspaceextra}{20pt}
\title{Chat2Scenario: Scenario Extraction From Dataset Through Utilization of Large Language Model\\
\thanks{\\
This work was supported by the National Key R\&D Program of China under Grant Nr. 2022YFE0117100, and by the FFG in the research project PECOP (FFG Projektnummer 893988), as part of the~\enquote{Bilateral Cooperation Austria - People's Republic of China / MOST 2nd Call} program. \textit{Corresponding author: Wenbo Xiao (wenbo.xiao@student.tugraz.at)}}
}

\author{\IEEEauthorblockN{Yongqi Zhao}
\IEEEauthorblockA{\textit{Institute of Automotive Engineering} \\
\textit{Graz University of Technology}\\
Graz, Austria}
\and
\IEEEauthorblockN{Wenbo Xiao}
\IEEEauthorblockA{\textit{Institute of Automotive Engineering} \\
\textit{Graz University of Technology}\\
Graz, Austria}
\and
\IEEEauthorblockN{Tomislav Mihalj}
\IEEEauthorblockA{\textit{Institute of Automotive Engineering} \\
\textit{Graz University of Technology}\\
Graz, Austria}
\and
\IEEEauthorblockN{Jia Hu}
\IEEEauthorblockA{\textit{College of Transportation Engineering} \\
\textit{Tongji University}\\
Shanghai, China}
\and
\IEEEauthorblockN{Arno Eichberger}
\IEEEauthorblockA{\textit{Institute of Automotive Engineering} \\
\textit{Graz University of Technology}\\
Graz, Austria}
}
    
\maketitle

\begin{abstract}
The advent of Large Language Models (LLM) provides new insights to validate Automated Driving Systems (ADS). In the herein-introduced work, a novel approach to extracting scenarios from naturalistic driving datasets is presented. A framework called Chat2Scenario is proposed leveraging the advanced Natural Language Processing (NLP) capabilities of LLM to understand and identify different driving scenarios. By inputting descriptive texts of driving conditions and specifying the criticality metric thresholds, the framework efficiently searches for desired scenarios and converts them into ASAM OpenSCENARIO\footnote{\url{https://www.asam.net/standards/detail/openscenario/}} and IPG CarMaker text files\footnote{\url{https://ipg-automotive.com/de/support/supportanfrage/faq/usage-of-user-inputs-from-a-file-in-a-maneuver-133/}}. This methodology streamlines the scenario extraction process and enhances efficiency. Simulations are executed to validate the efficiency of the approach. The framework is presented based on a user-friendly web app and is accessible via the following link:~\url{https://github.com/ftgTUGraz/Chat2Scenario}.
\end{abstract}


\section{Introduction}  \copyrightnotice
It has been proven that mileage-based on-road testing is not sufficient for the validation of ADS, as Automated Vehicles (AV) must be driven billions of miles to demonstrate their reliability~\cite{kalra2016driving}. To increase testing efficiency, a scenario-based method was proposed in project PEGASUS\footnote{\url{https://www.pegasusprojekt.de/en/home}} aiming to expose ADS in virtual driving environments derived from the real world. However, this approach heavily relies on the measurements of real-world traffic and high-fidelity simulation platforms. Adequate measurement data of real-world traffic ensures a reliable data source; simulators provide an efficient alternative to guarantee safety, as long as it is close to reality.

In recent years, there have been intensive investigations into contributions related to high-quality, cost-effective dataset provision (cf.~\cite{krajewski2018highd, zhang2023ad4che, bock2020ind, krajewski2020round, coifman2017critical, spannaus2021automatum, barmpounakis2020new, robicquet2016learning,9827305}) and the release of portable simulation platforms (cf.~\cite{dosovitskiy2017carla, ipg_automotive_carmaker, esmini}). However, the availability of portable and publicly accessible automation tools for reconstructing the measurements within these simulation platforms has been limited. Karunakaran et al.~\cite{karunakaran2022automatic} developed a tool for identifying and extracting lane change scenarios from LiDAR point clouds. Zhu et al.~\cite{zhu2023automatic} proposed a framework for extracting ADS disengagement scenarios from AV road testing data. Montanari et al.~\cite{9575441, OSC-Generator} created a tool to extract concrete scenarios from test vehicle bus communication data based on maneuvers. Zhang et al.~\cite{xinxin2020csg} introduced a toolkit to extract accident scenarios from traffic surveillance videos. However, these tools face several issues: 1) compatibility is limited to datasets that are either difficult to acquire at scale, not publicly accessible, or require extensive pre-processing; 2) the ability to extract only one type of scenario, resulting in a limited scope; 3) the inability to quantitatively evaluate the criticality of the generated scenarios; 4) the lack of open-source availability or user-friendly interfaces, making practical application laborious.

\begin{figure*}[htbp]
\centerline{\fbox{\includegraphics[width=0.95\textwidth]{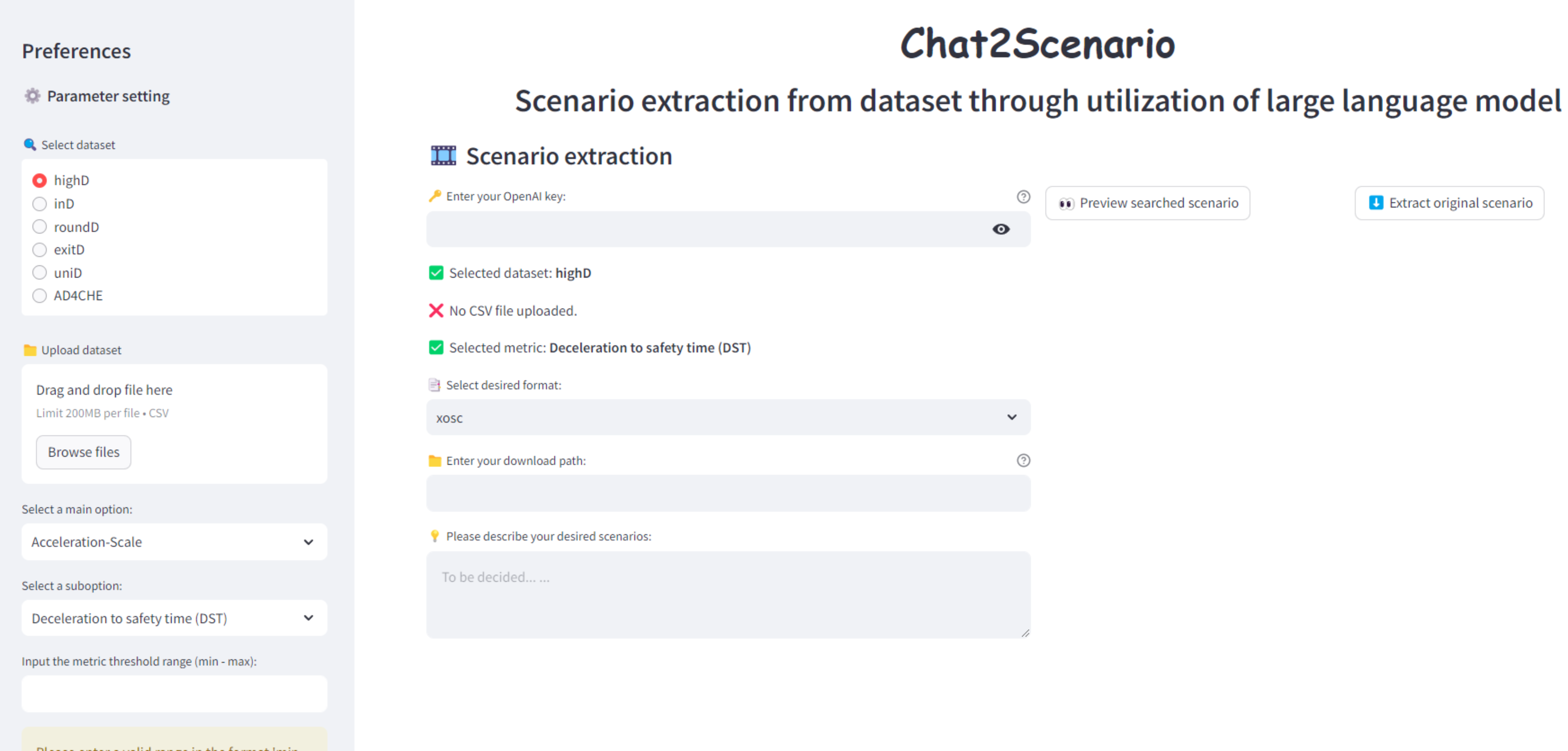}}}
\caption{Overview of Chat2Scenario web app.}
\label{fig:chat2scenario_web}
\end{figure*}

Based on the aforementioned issues, the Chat2Scenario framework is introduced. This framework incorporates the latest OpenAI LLM,~\textit{gpt-4-1106-preview}\footnote{\url{https://platform.openai.com/docs/models/gpt-4-and-gpt-4-turbo}}, to extract concrete scenarios from naturalistic driving datasets. In this work, the highD dataset (refer to~\cite{krajewski2018highd}), comprising vehicle trajectories on German highways collected via drones, is employed. This approach is advantageous not only because addresses the issue of limited data sources but also due to the minimal requirement for pre-processing. Datasets in the same format from various locations (e.g., intersection~\cite{bock2020ind}, roundabout~\cite{krajewski2020round}, highway exits~\cite{9827305}, university campus\footnote{\url{https://levelxdata.com/unid-dataset/}}) are also publicly available and can be potentially integrated into this tool.



Subsequently, the OpenAI LLM is leveraged to interpret scenarios described in natural language, thereby broadening the range of searchable scenario types. Next, several metrics are provided to quantify the criticality of generated scenarios. Finally, the tool is delivered as a user-friendly and practical web app (see Fig.~\ref{fig:chat2scenario_web}) to enhance usability. The contributions of this work are summarized as follows:

\begin{enumerate}
    \item The OpenAI LLM is utilized to enhance scenario searching efficiency and expand the searchable scenario types.
    \item Criticality metrics-based scenario filtering criterion is provided to promote the searching accuracy.
    \item A practical and shareable web app is released to connect the naturalistic driving dataset and the simulation platform for ADS validation.
\end{enumerate}

The presented framework would be useful to facilitate the process of the ADS function test. The outcome should also provide new insight for the ADS testing engineers to simplify the search and analysis of complex datasets.

\section{Terminology and Dataset Format}
\subsection{Definition of Activity and Event}
\begin{figure}[ht]
    \centering
    \begin{tikzpicture}
    \draw[very thick] (0,0) -- (6.5,0) node[anchor=north east]{};
    \draw[very thick, dashed, dash pattern=on 5pt off 15pt](0,-0.75)--(6.5,-0.75)node[anchor=north east]{};
    \draw[very thick] (0,-1.5) -- (6.5,-1.5) node[anchor=north east]{};
    \node (car1) at (1,-0.375){\includegraphics[width=1cm, angle=-90]{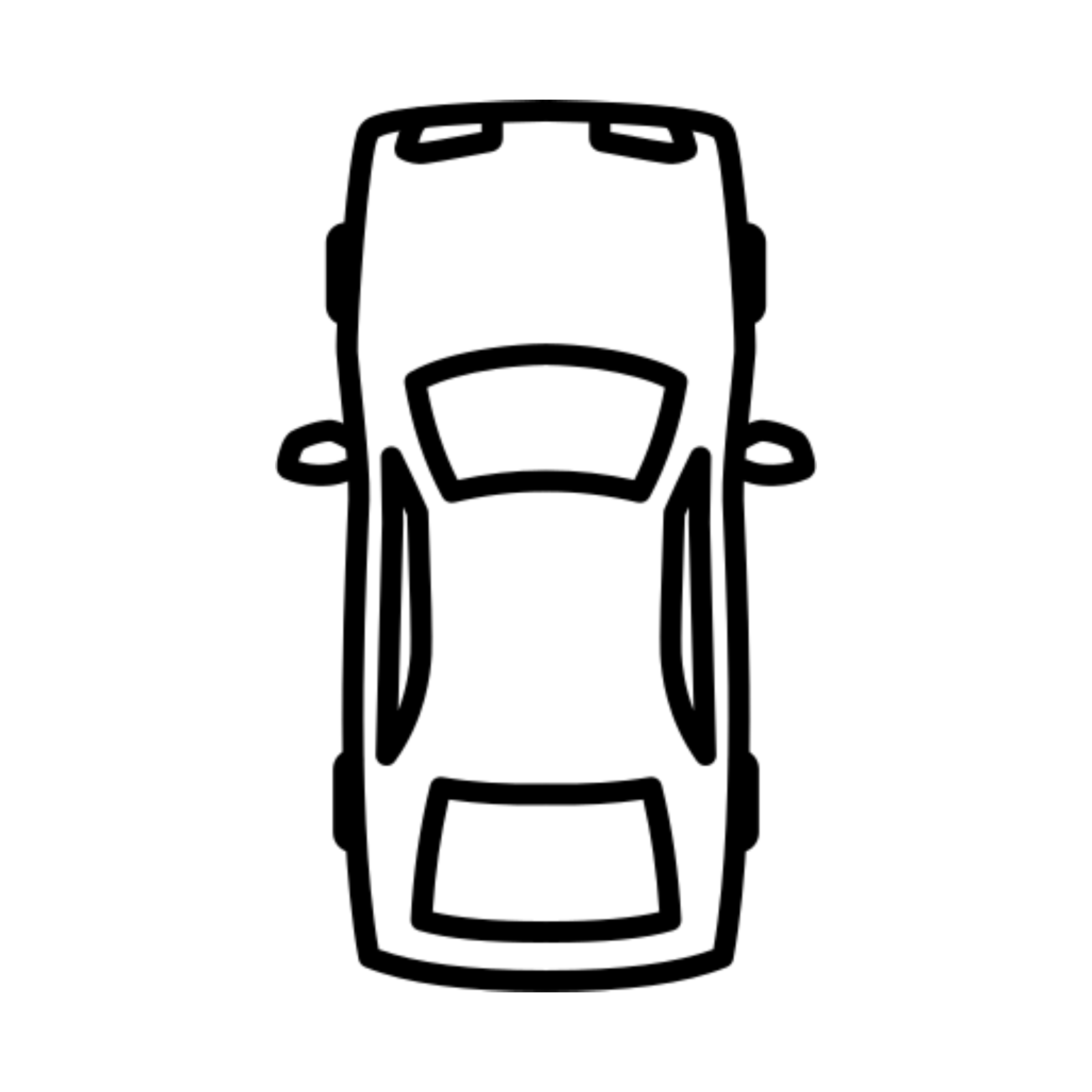}};
    \draw[-latex, blue](0.25,-0.375)--(2,-0.375)--(3.75,-1.15)--(4.75,-1.15);
    \draw[gray, dashed](0.25,-0.375)--(0.25, -2.1);
    \draw[gray, dashed](2,-0.375)--(2, -2.1);
    \draw[gray, dashed](3.75,-1.15)--(3.75, -2.1);
    \node[align=center] at (0.25,-2.2) {Event};
    \node[align=center] at (2,-2.2) {Event};
    \node[align=center] at (3.75,-2.2) {Event};
    \draw[<->, gray](0.25,-1.9)--(2,-1.9);
    \node[align=center] at (1.1,-1.75) {Activity};
    \draw[<->, gray](2,-1.9)--(3.75,-1.9);
    \node[align=center] at (2.8,-1.75) {Activity};
    \draw[<->, gray](3.75,-1.9)--(4.75,-1.9);
    \node[align=center] at (4.5,-1.75) {Activity};
    \draw[-latex, gray, ultra thick](0, 0.5)--(1.25, 0.5); 
    \node[align=center, gray]at(1.35, 0.5){\large x};
    \draw[-latex, gray, ultra thick](0, 0.5)--(0, -1.25); 
    \node[align=center, gray]at(-0.2, -1.2){\large y};
    \node[align=center, gray]at(-0.2, 0.5){\large 0};
    \node[align=center, red]at(6, -0.35){laneID: 1};
    \node[align=center, red]at(6, -1.15){laneID: 2};
    \end{tikzpicture}
    \caption{Visualization of~\textit{Event} and~\textit{Activity}: blue arrow represents the vehicle trajectory~\cite{elrofai2018scenario, krajewski2018highd}.}
    \label{fig:activity_illustration}
\end{figure}

An~\textit{Activity} is defined as~\textit{the minimal unit in a scenario's dynamics, representing the temporal progression of state variables, where its end signifies the commencement of the subsequent activity}~\cite{elrofai2018scenario}. An~\textit{Event} indicates~\textit{the time instant when a transition of state occurs, such that before and after an event, the state corresponds to two different activities}~\cite{elrofai2018scenario}. The concepts of~\textit{Event} and~\textit{Activity} are visualized in Fig.~\ref{fig:activity_illustration}.

\subsection{Dataset Format}
The highD dataset comprises multiple recordings, each encapsulated within a CSV file. Each file encompasses a suite of vehicle trajectories, providing comprehensive details such as the data frame, vehicle ID, position, velocity, acceleration, and the current lane ID for each respective trajectory, as depicted in Tab.~\ref{tab:highD_inf}. The global coordinate system's origin is positioned at the upper left corner, with the horizontal and vertical axes defined as the x-axis and y-axis, respectively. Lanes within this system are sequentially numbered starting from 1, as depicted in Fig.~\ref{fig:activity_illustration}. 

\begin{table}[htbp]
\caption{Available information in highD dataset~\cite{krajewski2018highd}}
\begin{center}
\begin{tabular}{|c|c|c|c|}
\hline
\textbf{Name} & \textbf{Unit} & \textbf{Name} & \textbf{Unit}\\
\hline
frame & [-] & xVelocity & [\si{\meter\per\second}]\\
\hline
id & [-] & yVelocity & [\si{\meter\per\second}]\\
\hline
x & [\si{\meter}] & xAcceleration & [m/$s^2$]\\
\hline
y & [\si{\meter}] & yAcceleration & [m/$s^2$]\\
\hline
width & [\si{\meter}] & laneId & [-]\\
\hline
height & [\si{\meter}] & ... & ...\\
\hline
\end{tabular}
\label{tab:highD_inf}
\end{center}
\end{table}

\begin{figure*}[htbp]
\centering
\resizebox{\textwidth}{!}{
\begin{tikzpicture}
\node (user) at (0,0) {\includegraphics[width=1cm]{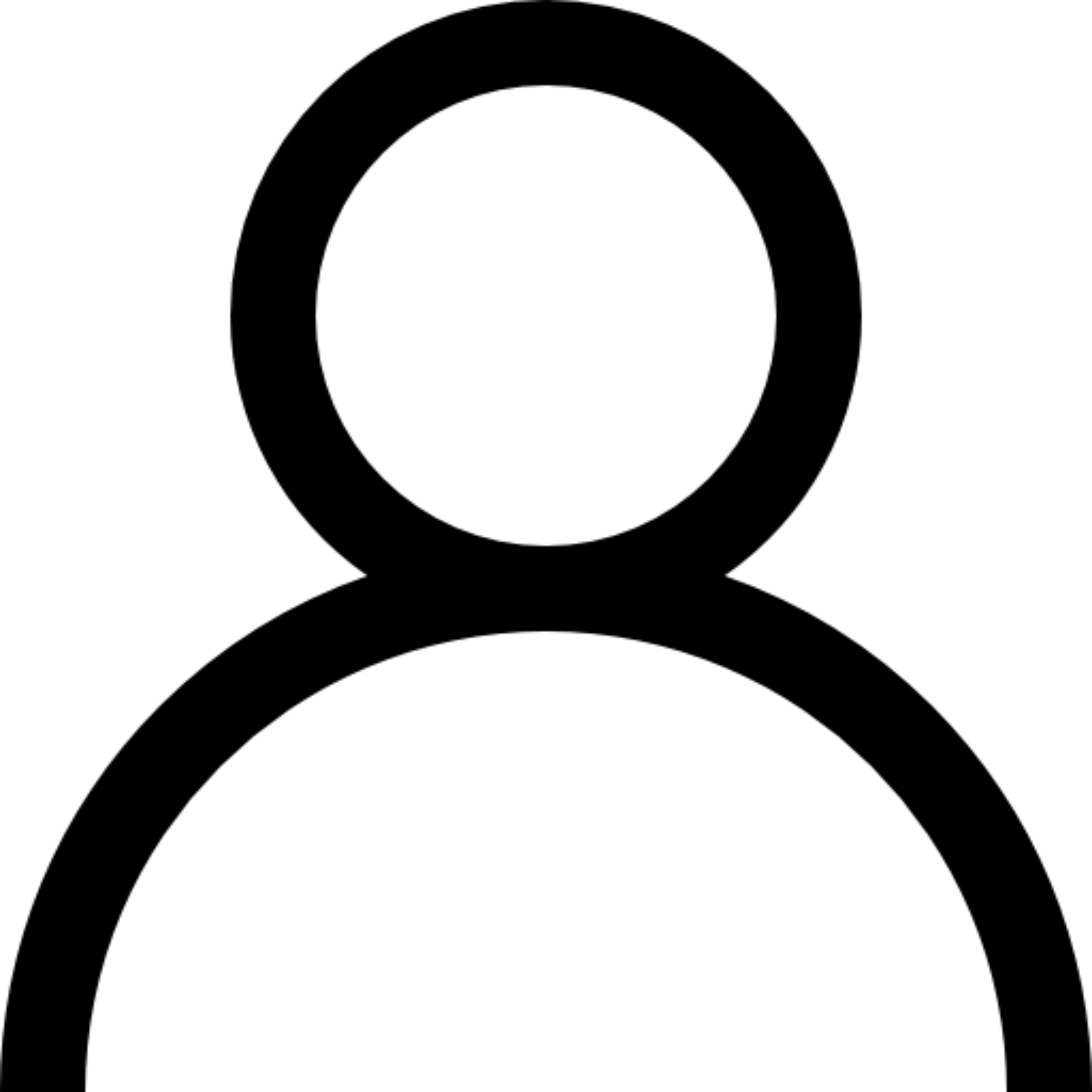}};
\node [below=0.05cm of user] (userText) {User};

\node[draw, rectangle, right=0.5cm of user, minimum width=3cm, minimum height=8cm, rounded corners](container){};

\node[draw, rectangle, dashed, minimum width=2.25cm, minimum height=2.5cm, rounded corners] at ([yshift=-1.5cm]container.north) (scenario) {};
\node at ([yshift=-0.9cm, xshift=-0.3cm]container.north)(car1){\includegraphics[width=.05\textwidth]{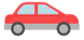}};
\node at ([yshift=-1.2cm, xshift=0.2cm]scenario.north)(car2){\includegraphics[width=.05\textwidth]{figures/car.png}};
\node[below=-1cm of scenario, align=center](label){Scenario \\visualization};

\node[draw, rectangle, dashed, minimum width=2.25cm, minimum height=4.5cm, rounded corners] at ([yshift=2.5cm]container.south) (paraSetup) {};
\node[rectangle, minimum width=2cm, minimum height=1cm, below=2mm of paraSetup.north, rounded corners, fill=gray!20](dataset){Dataset};
\node[rectangle, minimum width=2cm, minimum height=1cm, below=0.2cm of dataset, rounded corners, fill=gray!20, align=center](criticality){Criticality\\ metric \& \\threshold};
\node[rectangle, minimum width=2cm, minimum height=1cm, below=0.2cm of criticality, rounded corners, fill=gray!20, align=center](descriptive){Descriptive \\text of \\scenario};
\node[below=0cm of container]{Chat2Scenario web app};

\node[draw, rectangle, right=0.5cm of container, yshift=2.25cm, minimum width=5.5cm, minimum height=3.5cm, rounded corners](simFormatContainer){};
\node[rectangle, minimum width=2cm, minimum height=1cm, below=6mm of simFormatContainer.north, rounded corners, fill=gray!20, align=center, xshift=-1.5cm](openScenario){Open-\\SCENARIO};
\node[rectangle, minimum width=2cm, minimum height=1cm, below=3mm of openScenario, rounded corners, fill=gray!20, align=center](ipgCarMaker){IPG Car-\\Maker txt};
\node[rectangle, minimum width=2cm, minimum height=1cm, below=-4mm of openScenario, rounded corners, fill=gray!20, align=center, xshift=3cm](desiredSce){Selected\\scenario};
\draw[-latex] (desiredSce.west) -| ([xshift=0.5cm]openScenario.east) -- (openScenario.east);
\draw[-latex] (desiredSce.west) -| ([xshift=0.5cm]ipgCarMaker.east) -- (ipgCarMaker.east);
\node[below=0cm of simFormatContainer]{Simulatable format generation};

\node[draw, rectangle, right=0.5cm of simFormatContainer, minimum width=5.5cm, minimum height=3.5cm, rounded corners](cricAnalyContainer){};
\node[rectangle, minimum width=1.5cm, minimum height=1cm, below=12mm of cricAnalyContainer.north, rounded corners, fill=gray!20, align=center, xshift=-1.9cm](desiredSce2){Selected\\scenario};
\node[rectangle, minimum width=1.5cm, minimum height=1cm, right=3.5mm of desiredSce2, rounded corners, fill=gray!20, align=center](sceCandidate){Scenario\\pool};
\node[rectangle, minimum width=1.5cm, minimum height=1cm, below=4.5mm of cricAnalyContainer.north, rounded corners, fill=gray!20, align=center, xshift=1.8cm](criMetri){Criticality\\metric};
\node[rectangle, minimum width=1.5cm, minimum height=1cm, below=5mm of criMetri, rounded corners, fill=gray!20, align=center](criThresh){Criticality\\threshold};
\draw[-latex](criMetri.west) -|([yshift=2mm]sceCandidate.north) -- (sceCandidate.north);
\draw[-latex](criThresh.west) -| ([yshift=-2mm]sceCandidate.south) -- (sceCandidate.south);
\draw[-latex](sceCandidate.west) -- (desiredSce2.east);
\node[below=0cm of cricAnalyContainer]{Criticality analysis};

\node[draw, rectangle, below=1cm of simFormatContainer, minimum width=5.5cm, minimum height=3.5cm, rounded corners](sceUnderstandContainer){};
\node[rectangle, minimum width=1.5cm, minimum height=1cm, below=3mm of sceUnderstandContainer.north, rounded corners, fill=gray!20, align=center, xshift=-1.5cm](sceClassModel){Scenario\\classification\\model};
\node[rectangle, minimum width=1.5cm, minimum height=1cm, below=3mm of sceClassModel, rounded corners, fill=gray!20, align=center, xshift=1mm](descripTextSce){Descriptive\\text of\\scenario};
\node[rectangle, minimum width=1cm, minimum height=1cm, below=1.3cm of sceUnderstandContainer.north, xshift=0.65cm, rounded corners, fill=gray!20, align=center](prompt){Prompt};
\node[rectangle, minimum width=1cm, minimum height=1cm, right=0.3cm of prompt, rounded corners, fill=gray!20, align=center](llm){LLM};
\draw[-latex](sceClassModel.east) -| ([xshift=-2mm]prompt.west) -- (prompt.west);
\draw[-latex](descripTextSce.east) -| ([xshift=-2mm]prompt.west) -- (prompt.west);
\draw[-latex](prompt.east)  -- (llm.west);
\node[below=0cm of sceUnderstandContainer]{Scenario understanding};

\node[draw, rectangle, below=1cm of cricAnalyContainer, minimum width=5.5cm, minimum height=3.5cm, rounded corners](sceSearchContainer){};
\node[rectangle, minimum width=1.5cm, minimum height=1cm, below=4mm of sceSearchContainer.north, rounded corners, fill=gray!20, align=center, xshift=-1.5cm](dataset){Dataset};
\node[rectangle, minimum width=1.5cm, minimum height=1cm, below=0.7cm of dataset, rounded corners, fill=gray!20, align=center](llmRespo){LLM\\response};
\node[rectangle, minimum width=1.5cm, minimum height=1cm, right=0.5cm of dataset, rounded corners, fill=gray!20, align=center](actSceIden){Activity \&\\position\\analysis};
\node[rectangle, minimum width=1.5cm, minimum height=1cm, below=0.7cm of dataset, rounded corners, fill=gray!20, align=center, xshift=2.45cm](sceCandi){Scenario\\pool};
\draw[-latex](dataset)--(actSceIden);
\draw[-latex](llmRespo)--(sceCandi);
\draw[-latex](actSceIden.south) |-([yshift=2mm, xshift=-3mm]sceCandi.north)  -|([yshift=-5mm, xshift=-1.3cm]sceCandi.north);
\node[below=0cm of sceSearchContainer]{Scenario searching};

\draw[-latex, very thick](userText.south)|-([yshift=-1.15cm]userText.south)--([yshift=-2.3cm]container.west);
\draw[-latex, very thick]([yshift=-2.27cm]container.east)--(sceUnderstandContainer.west);
\draw[-latex, very thick](sceUnderstandContainer.east)--(sceSearchContainer.west);
\draw[-latex, very thick](sceSearchContainer.east)-|([xshift=5mm]sceSearchContainer.east)|-([xshift=5mm]cricAnalyContainer.east)--(cricAnalyContainer.east);
\draw[-latex, very thick](cricAnalyContainer.west)--(simFormatContainer.east);
\draw[-latex, very thick](simFormatContainer.west)--([yshift=2.25cm]container.east);
\draw[-latex, very thick]([yshift=2.3cm]container.west)-|([yshift=1.15cm]user.north)--(user.north);
\end{tikzpicture}
}
\caption{Schematic overview of the Chat2Scenario framework operations.}
\label{flowchart_chat2scenario}
\end{figure*}

\section{Methodology}
In Fig.~\ref{flowchart_chat2scenario}, it is illustrated the workflow of Chat2Scenario system. This process begins with the user uploading a dataset to the Chat2Scenario web app and entering a scenario description along with the criticality metric and threshold. LLM subsequently interprets the descriptive text of the scenario. Based on the LLM response, scenarios matching the requirements are added to the scenario pool, where their criticality is analyzed. Scenarios not meeting the threshold criteria are excluded. Finally, the selected scenarios are converted into ASAM OpenSCENARIO and IPG CarMaker text. These scenarios can be visualized via the web app. Each module in the flowchart is detailed in this section.

\subsection{Chat2Scenario Web App}
The shareable web app depicted in Fig.~\ref{fig:chat2scenario_web} is developed using the Python programming language and leverages the~\textit{Streamlit}\footnote{\url{https://docs.streamlit.io/library}} library. It offers a user-friendly interface that enables users to configure parameters and visually explore selected scenarios with ease.

\begin{figure*}[htbp]
    \centering
    \resizebox{\textwidth}{!}{
    \begin{tikzpicture}[
        block/.style={rectangle, text width=2cm, align=center, minimum height=1cm, fill=gray!20},
        group/.style={rectangle, draw, thick, inner sep=10pt, outer sep=0pt}
    ]
      \node[block] (keepVelocity) {Keep velocity};
      \node[block, right=0.5cm of keepVelocity] (acceleration) {Acceleration};
      \node[block, right=0.5cm of acceleration] (deceleration) {Deceleration};

      \node[block, right=1.5cm of deceleration] (followLane) {Follow lane};
      \node[block, right=0.5cm of followLane] (laneChangeLeft) {Lane change left};
      \node[block, right=0.5cm of laneChangeLeft] (laneChangeRight) {Lane change right};
    
      \node[group, fit=(keepVelocity) (acceleration) (deceleration), above=-11.5mm of acceleration] (longitudinal) {};
      \node[group, fit=(followLane) (laneChangeLeft) (laneChangeRight), above=-11.5mm of laneChangeLeft] (lateral) {};

      \node[above=-5mm of longitudinal.north] (longitudinalLabel) {Longitudinal Activity};
      \node[above=-5mm of lateral.north] (lateralLabel) {Lateral Activity};

      \node[group, very thick, fit=(longitudinal) (lateral), above=-13.5mm of acceleration, xshift=46mm] () {};
      \node[above=5.5mm of acceleration, xshift=44mm] (egoTgtActivity) {Ego/Target Vehicle Activity};

      \node[block, text width=3.35cm, above=-37mm of keepVelocity, xshift=7mm](front){Front};
      \node[block, text width=3.35cm, right=0.5cm of front](behind){Behind};
      \node[group, fit=(front)(behind), above=-12mm of front, xshift=19.75mm] (sameLane) {};
      \node[above=1mm of front, xshift=20mm] (sameLaneLabel) {Same Lane};

      \node[block, text width=3.35cm, right=1.5cm of behind](leftAdjacent){Left adjacent lane};
      \node[block, text width=3.35cm, right=0.5cm of leftAdjacent](rightAdjacent){Right adjacent lane};
      \node[group, fit=(leftAdjacent)(rightAdjacent), above=-12mm of leftAdjacent, xshift=19.75mm] (adjacentLane) {};
      \node[above=0mm of leftAdjacent, xshift=20mm](adjacentLaneLabel){Adjacent Lane};

      \node[block, text width=8cm, below=0.9cm of front, xshift=23mm](nextLeftAdjacent){Lane next to left adjacent lane};
      \node[block, text width=8cm, right=0.5cm of nextLeftAdjacent](nextRightAdjacent){Lane next to right adjacent lane};
      \node[group, fit=(nextLeftAdjacent)(nextRightAdjacent), above=-12mm of nextLeftAdjacent, xshift=43mm](laneNextToAdjacentLane){};
      \node[above=0mm of nextLeftAdjacent, xshift=45mm](laneNextToAdjacentLaneLabel){Lane Next to Adjacent Lane};

      \node[group, very thick, fit=(sameLane)(adjacentLane)(laneNextToAdjacentLane), above=-33mm of front, xshift=66.5mm](targetVehiclePosition){};
      \node[above=0mm of sameLane, xshift=45mm](targetVehiclePositionLabel){Target Vehicle Position};
      
    \end{tikzpicture}
    }
    \caption{Scenario classification model for highD traffic.}
    \label{fig:scenario_classification_model}
\end{figure*}
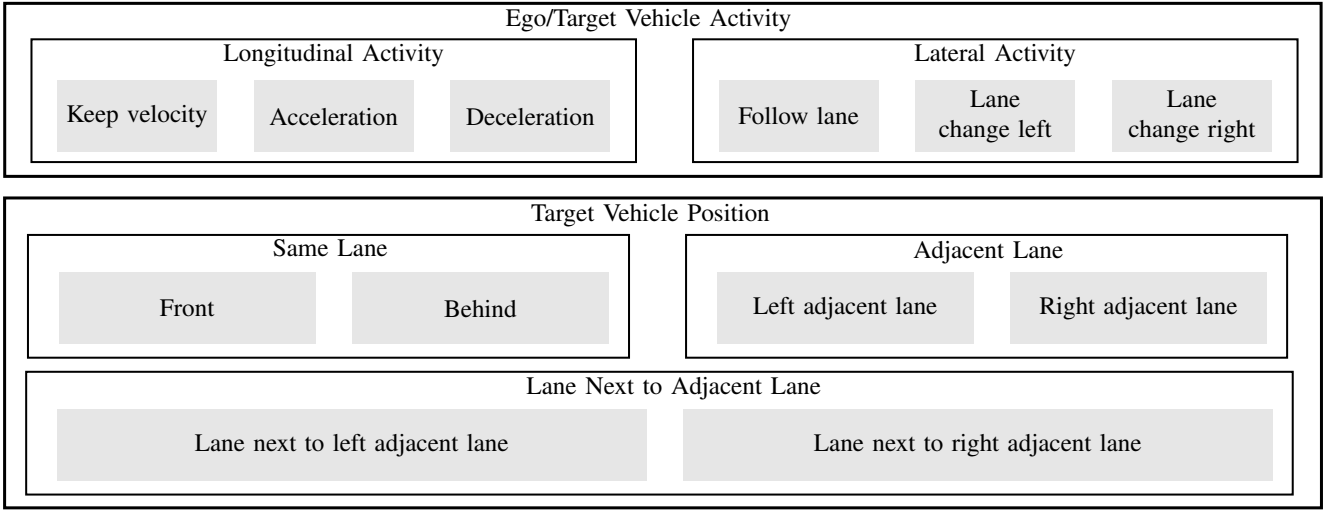

\subsection{Scenario Understanding}
In the scenario understanding module, the NLP capabilities of LLM are utilized to identify and categorize the dynamic behaviors and positions of vehicles within driving scenarios. A key factor in this process is the strategic prompt engineering, which directs the LLM to generate precise and relevant responses. In this study, the prompts provided to the LLM integrate a scenario classification model. This integration allows the LLM to interpret scenario narratives and accurately align semantic labels with a well-structured framework, transforming unstructured text into structured scenario data. Further details on the scenario classification model and the interface and the prompt engineering are elaborated in subsequent sections.

\subsubsection{Scenario Classification Model}
In Fig.~\ref{fig:scenario_classification_model}, the classification model for highway traffic scenarios is illustrated. This model categorizes the information into two primary sections: vehicle activity and target vehicle's position w.r.t. the ego vehicle. Vehicle activity is subdivided into longitudinal and lateral activities. Longitudinal activity pertains to velocity, with three possible states:~\textit{Keep velocity},~\textit{Acceleration}, or~\textit{Deceleration}~\cite{hartjen2019classification}. Lateral activity relates to the vehicle's interaction with traffic lanes, comprising~\textit{Follow lane},~\textit{Lane change left}, or~\textit{Lane change right}.

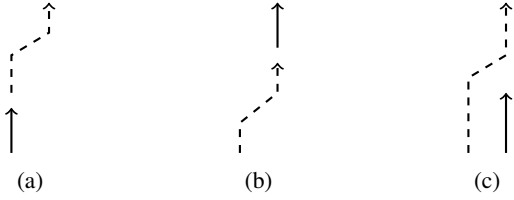
\begin{figure}[ht]
    \centering
    \begin{subfigure}[b]{0.3\columnwidth}
        \centering
        \begin{tikzpicture}
            \draw[thick,->] (0,0) -- (0, 0.6);
            \draw[thick,->,dashed] (0,0.8) to (0,1.3) to (0.5,1.6) to (0.5,2);
        \end{tikzpicture}
        \caption{}
        \label{fig:error_1}
    \end{subfigure}
    \hfill
    \begin{subfigure}[b]{0.3\columnwidth}
        \centering
        \begin{tikzpicture}
            \draw[thick,->,dashed] (0,0) to (0, 0.4) to (0.5,0.8) to (0.5,1.2);
            \draw[thick,->] (0.5,1.4) -- (0.5,2);
        \end{tikzpicture}
        \caption{}
        \label{fig:error_2}
    \end{subfigure}
    \hfill
    \begin{subfigure}[b]{0.3\columnwidth}
        \centering
        \begin{tikzpicture}
            \draw[thick,->,dashed] (0,0) to (0, 1) to (0.5,1.3) to (0.5,2);
            \draw[thick,->] (0.5,0) -- (0.5,0.8);
        \end{tikzpicture}
        \caption{}
        \label{fig:desired_cut_in}
    \end{subfigure}
    \caption{Visualization of scenarios: dashed and solid lines represent target and ego vehicle trajectories respectively.}
    \label{fig:main}
\end{figure}

The relative position of the target vehicle concerning the ego vehicle is crucial for resolving ambiguities in scenarios. For instance, consider an unspecified scenario where~\textit{the ego vehicle follows the lane, and a target vehicle changes lane to the right}. This situation could correspond to any scenarios depicted in Fig.~\ref{fig:error_1}~-~\ref{fig:desired_cut_in}. However, if it is specified that~\textit{the target vehicle begins in the left adjacent lane and ends up in front of the ego vehicle within the same lane}, only the scenario in Fig.~\ref{fig:desired_cut_in} satisfies these conditions.

Target vehicles position w.r.t. ego vehicle is categorized as being in the~\textit{Same Lane},~\textit{Adjacent Lane}, or~\textit{Lane Next to Adjacent Lane}. The presence of vehicles in the same or adjacent lanes is significant for the decision-making of the AV and is thus included~\cite{9773877}. Vehicles in the lane next to the adjacent lane are also considered due to their potential to merge into the ego vehicle's lane~\cite{9773877}. Vehicles in other lanes are excluded from consideration as they are unlikely to interact with the ego vehicle. Regarding the~\textit{Same Lane}, target vehicles are either~\textit{Behind} or in~\textit{Front} of the ego vehicle. As for the~\textit{Adjacent lane}, the target vehicles are situated in the~\textit{Left} or~\textit{Right adjacent lane}. The categorization is analogous for vehicles in the~\textit{Lane Next to Adjacent Lane}.

\subsubsection{Prompt Engineering of LLM}

\begin{figure*}[htbp]
\centering
\begin{minipage}[c]{\textwidth}
\begin{minipage}[c]{.035\textwidth}
    \subcaption{} 
    \label{fig:personaBoxSub}
\end{minipage}%
\begin{minipage}[c]{.8\textwidth}
\begin{tikzpicture}
\node[draw, minimum width=8cm, minimum height=2cm, align=left, text width=17cm, fill=gray!20] (personaBox) {System, you are an AI trained to understand and classify driving scenarios based on specific frameworks. Your task is to analyze the following driving scenario and classify the behavior of both the ego vehicle and the target vehicle according to the given classification framework. Please follow the framework strictly and provide precise and clear classifications. The framework is as follows:~\textbf{scenario\_classification\_model}};
\end{tikzpicture}
\end{minipage}
\end{minipage}

\vspace{5pt}

\begin{minipage}[c]{\textwidth}
\begin{minipage}[c]{.035\textwidth}
    \subcaption{} 
    \label{fig:sceDespBoxSub}
\end{minipage}%
\begin{minipage}[c]{.8\textwidth}
\begin{tikzpicture}
\node[draw, minimum width=8cm, minimum height=0.5cm, align=left, text width=17cm, fill=gray!20](sceDespBox){Scenario Description:~\textbf{descriptive\_text\_of\_scenario}};
\end{tikzpicture}
\end{minipage}
\end{minipage}

\vspace{5pt}

\begin{minipage}[c]{\textwidth}
\begin{minipage}[c]{.035\textwidth}
    \subcaption{} 
    \label{fig:taskDetailSub}
\end{minipage}%
\begin{minipage}[c]{.8\textwidth}
\begin{tikzpicture}
\node[draw, minimum width=8cm, minimum height=0.5cm, align=left, text width=17cm, fill=gray!20, below=0.2cm of sceDespBox](taskDetailBox){Provide a detailed classification for both the ego vehicle and the target vehicle(s). The response should be formatted exactly as shown in this structure:\\
\{\\~~Ego Vehicle:~\{Ego longitudinal activity: ['Your Classification'], Ego lateral activity: ['Your Classification']\},\\
~~Target Vehicle \#1:\\
~~\{\\
~~~~~~~~Target start position: \{'Your Classification': ['Your Classification']\},\\
~~~~~~~~Target end position: \{'Your Classification': ['Your Classification']\},\\
~~~~~~~~Target behavior: \{target longitudinal activity: ['Your Classification'],\\
~~~~~~~~~~~~~~~~~~~~~~~~~~~~~~target lateral activity: ['Your Classification']'\}\\
~~\}
\\
~~Target Vehicle \#2:\\
~~\{\\
~~~~~~......\\
~~~~~~......\\
~~\}
\\
\}};
\end{tikzpicture}
\end{minipage}
\end{minipage}

\vspace{5pt}

\begin{minipage}[c]{\textwidth}
\begin{minipage}[c]{.035\textwidth}
    \subcaption{} 
    \label{fig:exampleSub}
\end{minipage}%
\begin{minipage}[c]{.8\textwidth}
\begin{tikzpicture}
\node[draw, minimum width=8cm, minimum height=2cm, align=left, text width=17cm, fill=gray!20, below=0.2cm of sceDespBox] (exampleBox) {Example: If an ego vehicle is maintaining speed and following its lane, while another vehicle is initially in the left adjacent lane and is accelerating, then changing lanes to the right; finally driving on the front of ego vehicle, the classification would be:\\
\{\\~~Ego Vehicle:~\{Ego longitudinal activity: ['keep velocity'], Ego lateral activity: ['follow lane']\},\\
~~Target Vehicle:\\
~~\{\\
~~~~~~~~Target start position: \{'adjacent lane': ['left adjacent lane']\},\\
~~~~~~~~Target end position: \{'same lane': ['front']\},\\
~~~~~~~~Target behavior: \{target longitudinal activity: ['acceleration'],\\
~~~~~~~~~~~~~~~~~~~~~~~~~~~~~~target lateral activity: ['lane change right']'\}\\
~~\}
};
\end{tikzpicture}
\end{minipage}
\end{minipage}

\vspace{5pt}

\begin{minipage}[c]{\textwidth}
\begin{minipage}[c]{.035\textwidth}
    \subcaption{} 
    \label{fig:endSub}
\end{minipage}%
\begin{minipage}[c]{.8\textwidth}
\begin{tikzpicture}
\node[draw, minimum width=8cm, minimum height=0.5cm, align=left, text width=17cm, fill=gray!20, below=0.2cm of exampleBox](endBox){Remember to analyze carefully and provide the classification as per the structure given above.};
\end{tikzpicture}
\end{minipage}
\end{minipage}
\caption{Description prompt submitted to LLM.}
\label{fig:prompt_LLM}
\end{figure*}

Prompt engineering refers to the strategic formulation of input queries to effectively guide the LLM's responses, optimizing for more accurate, relevant, and useful outputs. Informed by the six strategies from OpenAI's prompt engineering guide\footnote{\url{https://platform.openai.com/docs/guides/prompt-engineering}}, a structured prompt for LLM optimization is proposed, as depicted in Figure~\ref{fig:prompt_LLM}. The prompt consists of five segments. The first segment, Fig.~\ref{fig:personaBoxSub}, delineates the role of the LLM as an advanced AI tool for scenario analysis, specially tasked with interpreting driving scenarios following a pre-established classification model shown in Fig.~\ref{fig:scenario_classification_model}. This structured approach is designed to minimize response variability, thereby simplifying the subsequent processing. 

In Fig.~\ref{fig:sceDespBoxSub}, the user is required to provide a detailed description of the driving scenario as~'\textbf{descriptive\_text\_of\_scenario}'. This input serves as the contextual basis for the LLM's analytical extraction.  

The details of the task are articulated in Fig.~\ref{fig:taskDetailSub}, which sets forth the expected structured response from the LLM. It is necessary to use curly braces '\{\}' and square brackets '[]' to systematically encapsulate the identified attributes of the scenario. In instances involving several target vehicles, the response employs a numbering system with '\#' and a sequential numeral to organize the data. This structured format is imperative for the automated parsing of scenario data, thus optimizing programming efficiency.

An application of this framework is demonstrated in Fig.~\ref{fig:exampleSub}, providing an example of how the LLM should process a specific driving situation. It depicts an instance in which the ego vehicle maintains its velocity and lane position while another vehicle executes lane changes and accelerations. The subsequent subprompt in Fig.~\ref{fig:endSub} reinforces the analytical task, guaranteeing precision and uniformity in the results.

\subsection{Scenario Searching}
The primary objective of scenario searching is to evaluate the congruence between vehicle trajectories in the datasets and the LLM's responses. This congruence assessment hinges on the identified activities of both the ego and target vehicles, as well as their relative positional relationships. The methodologies are expounded upon in the following sections.

\subsubsection{Activity Identification}
The distinction between sub-categories of longitudinal activities~\(A_{\text{lon}}\) depends on the comparison of longitudinal acceleration~\(a_{\text{lon}}\) with a predefined acceleration threshold~\(a_\text{lon}^\text{thr}\)~\cite{BOKARE20174733}. This comparison must satisfy the following equation:
\begin{equation}
A_\text{lon}(a_\text{lon}) = 
\begin{cases} 
\textit{Deceleration,} & a_\text{lon} < -a_\text{lon}^\text{thr}, \\
\textit{Acceleration,} & a_\text{lon} > a_\text{lon}^\text{thr}, \\
\textit{Keep velocity,} & \text{otherwise.} 
\end{cases}
\end{equation}

The variation in lane ID, denoted as ($\Delta$\(L\)), indicates lateral activities~\(A_\text{lat}\) and helps in identifying lane changes. Additionally, the vehicle's longitudinal velocity~(\(v_\text{lon}\)), in relation to the x-axis, determines the direction of the change, whether left or right. This is quantified by the following equation:
\begin{equation}
A_\text{lat} = \\
\begin{cases}
\textit{Follow lane}, & \Delta L = 0, \\
\textit{Lane change right}, & (\Delta L > 0 \text{ and } v_\text{lon} > 0) \text{ or } \\
& (\Delta L < 0 \text{ and } v_\text{lon} < 0),\\
\textit{Lane change left}, & (\Delta L < 0~\text{and}~v_\text{lon} > 0)~\text{or} \\&(\Delta L > 0~\text{and}~v_\text{lon} < 0).
\end{cases}
\end{equation}

\subsubsection{Position Identification}
The relative position of the target vehicle w.r.t. the ego vehicle, denoted as \(P_\text{tgt}^\text{ego}\), can be determined by the absolute value of the lane ID difference between the target and ego vehicles, given by $\|$\(\Delta L_\text{ego}^\text{tgt}\)$\|$, which is the modulus of the difference in lane IDs, $\|$\(L_\text{tgt} - L_\text{ego}\)$\|$, and the position difference along the x-axis, $\Delta x$.

For situations where \(v_\text{lon}\) $<$ 0, the position \(P_\text{tgt}^\text{ego}\) should satisfy the following conditions:

\begin{equation}
P_\text{tgt}^\text{ego} =
\left\{
\begin{array}{ll}
\textit{Front}, & \|\Delta L_\text{ego}^\text{tgt}\| = 0 \text{ and } \Delta x < 0, \\
\textit{Behind}, & \|\Delta L_\text{ego}^\text{tgt}\| = 0 \text{ and } \Delta x > 0, \\
\textit{Left adjacent lane}, & \Delta L_\text{ego}^\text{tgt} = 1, \\
\textit{Right adjacent lane}, & \Delta L_\text{ego}^\text{tgt} = -1, \\
\begin{array}{@{}l}
\textit{Lane next to the} \\
\textit{left adjacent lane},
\end{array} & \Delta L_\text{ego}^\text{tgt} = 2, \\
\begin{array}{@{}l}
\textit{Lane next to the} \\
\textit{right adjacent lane},
\end{array} & \Delta L_\text{ego}^\text{tgt} = -2. \\
\end{array}
\right.
\end{equation}

Similarly, when \(v_\text{lon}\) $>$ 0, the position \(P_\text{tgt}^\text{ego}\) is defined by:
\begin{equation}
P_\text{tgt}^\text{ego} =
\left\{
\begin{array}{ll}
\textit{Front}, & \|\Delta L_\text{ego}^\text{tgt}\| = 0 \text{ and } \Delta x > 0, \\
\textit{Behind}, & \|\Delta L_\text{ego}^\text{tgt}\| = 0 \text{ and } \Delta x < 0, \\
\textit{Left adjacent lane}, & \Delta L_\text{ego}^\text{tgt} = -1, \\
\textit{Right adjacent lane}, & \Delta L_\text{ego}^\text{tgt} = 1, \\
\begin{array}{@{}l}
\textit{Lane next to the} \\
\textit{left adjacent lane},
\end{array} & \Delta L_\text{ego}^\text{tgt} = -2, \\
\begin{array}{@{}l}
\textit{Lane next to the} \\
\textit{right adjacent lane},
\end{array} & \Delta L_\text{ego}^\text{tgt} = 2. \\
\end{array}
\right.
\end{equation}

\subsection{Criticality Analysis}

The scenario pool may encompass a multitude of scenarios that correspond to the input descriptive text. To quantitatively ascertain the scenarios most pertinent to the testing task, their criticalities must be evaluated. Consequently, this study incorporates some metrics summarized by Westhofen et al.~\cite{westhofen2023criticality} for assessing scenario criticality. 

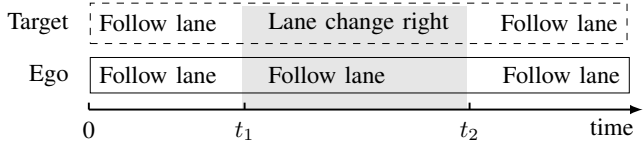
\begin{figure}[htbp]
    \centering
    \resizebox{\columnwidth}{!}{
    \begin{tikzpicture}
    \node[rectangle, fill=gray!20, minimum width=3.25cm, minimum height=1.5cm]at(3.34,0.75)(){};
    \draw[-latex, thick] (-0.5,0) -- (7.5,0) node[anchor=north east] {time};
    
    \node[above]at(0.5,1)(tgtFL){Follow lane};
    \node[right=0.5cm of tgtFL](tgtLcr){Lane change right};
    \node[right=0.5cm of tgtLcr](tgtFL2){Follow lane};
    
    \node[above] at (0.5,0.25)(egoFL){Follow lane};
    \node[right=0.5cm of egoFL](egoFL2){Follow lane};
    \node[right=1.45cm of egoFL2](egoFL3){Follow lane};
    
    \draw[thick](-0.5, 0)--(-0.5, 0.1);
    \draw[thick](1.75, 0)--(1.75, 0.1);
    \draw[thick](5, 0)--(5, 0.1);

    \node[below]at(-0.5, -0.05){0};
    \node[below]at(1.75, -0.05){$t_1$};
    \node[below]at(5, -0.05){$t_2$};

    \node[draw, rectangle, dashed, fit=(tgtFL)(tgtLcr)(tgtFL2),inner sep=0.01cm]{};
    \node[draw, rectangle, fit=(egoFL)(egoFL2)(egoFL3),inner sep=0.01cm]{};

    \node[left=0.2cm of tgtFL](tgt){Target};
    \node[left=0.2cm of egoFL](ego){Ego};
    
    \end{tikzpicture}
    }
    \caption{Illustration of criticality analysis within a scenario: the gray area denotes the segment under analysis.}
    \label{fig:criticality_analysis_time}
\end{figure}

A further consideration is determining which scenes within a scenario warrant criticality analysis. In this work, criticality is computed exclusively for the scenario segments where both the ego and target vehicles' activities and their relative positions are satisfied. For example, in the scenario depicted in Fig.~\ref{fig:desired_cut_in}, criticality is assessed from $t_1$ to $t_2$, as shown in Fig.~\ref{fig:criticality_analysis_time}. The criticality metrics utilized in Chat2Scenario are cataloged in Table~\ref{tab:criticality_metric}. 

\begin{table}[htbp]
\caption{Criticality Metric in Chat2Scenario~\cite{westhofen2023criticality}}
\begin{center}
\begin{tabular}{|c|c|}
\hline
\textbf{Metric Category} & \textbf{Metric Name}\\
\hline
\multirow{4}{*}{Acceleration-Scale} & Deceleration to safety time (DST)\\
& Required longitudinal acceleration (RLongA)\\
& Required lateral acceleration (RLatA)\\
& Required acceleration (RA)\\
\hline
\multirow{2}{*}{Distance-Scale} & Proportion of stopping distance (PSD)\\
& Distance headway (DHW)\\
\hline
\multirow{2}{*}{Jerk-Scale} & Longitudinal jerk (LongJ)\\
& Lateral jerk (LatJ)\\
\hline
\multirow{11}{*}{Time-Scale} & Encroachment time (ET)\\
& Post-encroachment time (PET)\\
& Time to collision (TTC)\\
& Potential time to collision (PTTC)\\
& Time exposed TTC (TET)\\
& Time integrated TTC (TIT)\\
& Time to closest encounter (TTCE)\\
& Time to brake (TTB)\\
& Time to kickdown (TTK)\\
& Time to steer (TTS)\\
& Time headway (THW)\\
\hline
Velocity-Scale & $\Delta v$\\
\hline
\end{tabular}
\label{tab:criticality_metric}
\end{center}
\end{table}

\subsection{Simulatable Format Generation}
The Chat2Scenario platform facilitates the generation of scenarios in two formats: ASAM OpenSCENARIO and IPG CarMaker text. 
\subsubsection{ASAM OpenSCENARIO}
The OpenSCENARIO files, generated using the~\textit{scenariogeneration}\footnote{\url{https://github.com/pyoscx/scenariogeneration}} Python package, are compatible with multiple simulators, including Esmini~\cite{esmini} and CARLA~\cite{dosovitskiy2017carla}. The XML schema is utilized for defining the scenarios in OpenSCENARIO. In this work, scenarios are reconstructed in Esmini based on vehicular trajectories. This process involves specifying a Vertex for each timestamp, illustrated in Fig.~\ref{fig:xosc}. For each Vertex, the vehicle's WorldPosition is delineated, with coordinates ($x, y, z$) derived directly from the dataset. Regarding the attitude, the heading angle ($h$) is constrained to 0 or $\pi$ radians, reflecting the road's alignment with the x-axis. Both pitch ($p$) and roll ($r$) angles are set to 0 radians by default. 

\begin{figure}[htbp]
\centering
\resizebox{0.95\columnwidth}{!}{
\begin{tikzpicture}
\node[draw, minimum width=4cm, minimum height=2cm, align=left, text width=8.5cm] (xoscBox)
{
\textcolor{orange}{$<$Vertex} \textcolor{blue}{time=}\textcolor{orange}{"0.0"$>$}\\
~~~~\textcolor{orange}{$<$Position$>$}\\
~~~~~~~~\textcolor{orange}{$<$WorldPosition} \textcolor{blue}{x=}\textcolor{orange}{"389.16"} \textcolor{blue}{y=}\textcolor{orange}{"-14.27"} \textcolor{blue}{z=}\textcolor{orange}{"0.0"} \textcolor{blue}{h=}\textcolor{orange}{"0.0"} \textcolor{blue}{p=}\textcolor{orange}{"0.0"} \textcolor{blue}{r=}\textcolor{orange}{"0.0"/$>$}\\
~~~~\textcolor{orange}{$<$/Position}$>$\\
\textcolor{orange}{$<$/Vertex}$>$
};
\end{tikzpicture}
}
\caption{An exemplary vertex in OpenSCENARIO.}
\label{fig:xosc}
\end{figure}

\subsubsection{IPG CarMaker Text}
The IPG CarMaker text format is tailored to incorporate real-world measurements of vehicle maneuvers into the CarMaker simulation environment. As depicted in Fig.~\ref{fig:ipg_CM_text}, the format begins with a timestamp in the first column. Successive columns record the vehicle's global position coordinates - longitudinal ($x$) and lateral ($y$) positions. The initial row captures the vehicle's position at the scene's start, with a timestamp of zero and \enquote{162} as the vehicle's identifier. In scenarios with multiple vehicles, their data are adjacently aligned, employing unique identifiers for each. Notably, this format is applied solely to non-ego vehicles. Regarding the ego vehicle, the trajectory is defined on the road network through the edition of UserPath.Nodes. More details are available in the IPGRoad document~\cite{IPGCarMaker2022}.

\begin{figure}[htbp]
\centering
\begin{tikzpicture}
\node[draw, minimum width=1cm, minimum height=0.5cm, align=center, text width=5cm](textBox)
{
\begin{tabular}{c c c c}
\textcolor{blue}{\#time,} & \textcolor{blue}{x\_162,} & \textcolor{blue}{y\_162,} & \textcolor{blue}{...}\\
\textcolor{blue}{0.0,} & \textcolor{blue}{389.16,} & \textcolor{blue}{-14.27,} & \textcolor{blue}{...}\\
\textcolor{blue}{...} & \textcolor{blue}{...} & \textcolor{blue}{...} & \textcolor{blue}{...}\\
\end{tabular}
};
\end{tikzpicture}
\caption{Illustration of IPG CarMaker text format.}
\label{fig:ipg_CM_text}
\end{figure}

\section{Result and Discussion}

\subsection{Qualitative Evaluation}

\begin{table*}[htbp]
\caption{Exemplary extracted scenarios through the utilization of Chat2Scenario}
\begin{center}
\begin{tabular}{|c|m{2.7cm}|m{6cm}|m{6cm}|}
\hline
\textbf{}&\centering \textbf{Descriptive Text}& \centering \textbf{Extracted Scenario (Esmini)}& \multicolumn{1}{c|}{\textbf{Extracted Scenario (CarMaker)}}\\
\hline
\rotatebox[origin=c]{90}{Following} & The ego vehicle follows the lane and decelerates. Target vehicle \#1, which is in front of the ego vehicle in the same lane, also decelerates. & \includegraphics[width=6cm, height=3.7cm, trim= 0 0 0 -2mm, keepaspectratio]{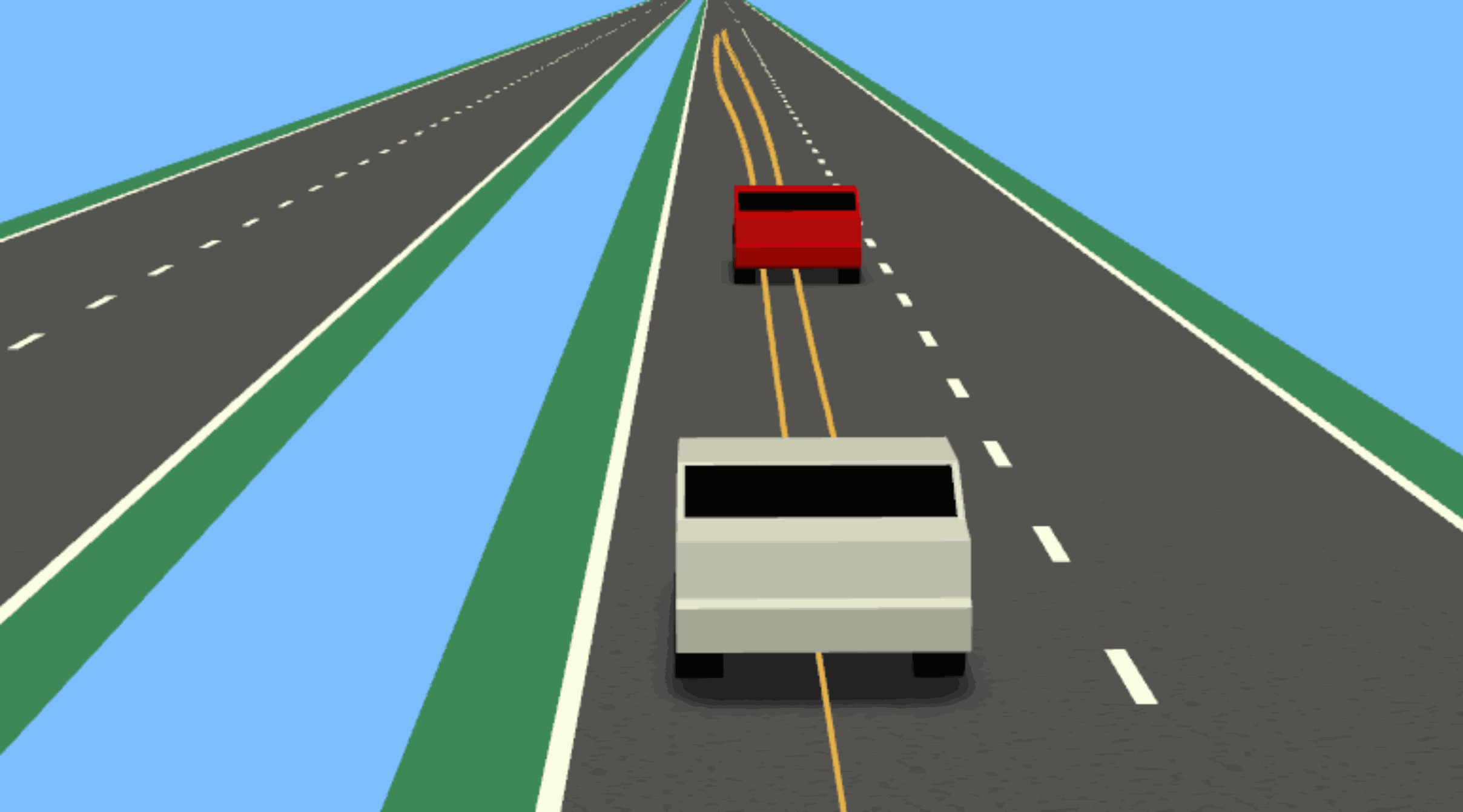} & \includegraphics[width=6cm, height=3.7cm, trim= 0 -6mm 0 -8mm]{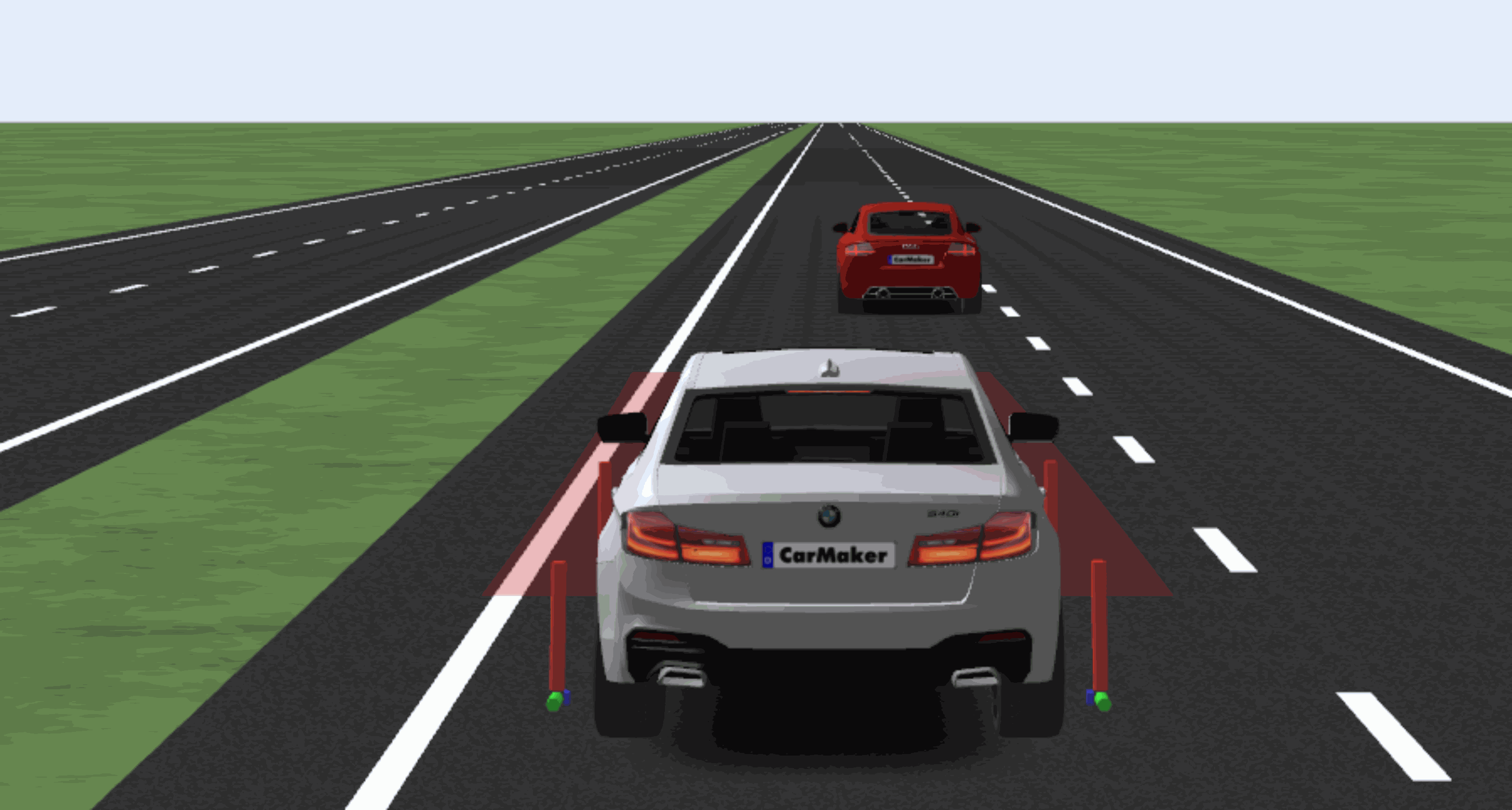} \\
\hline
\rotatebox[origin=c]{90}{Cut-in} & The ego vehicle maintains its lane and velocity. Initially, Target Vehicle \#1 is driving in the left adjacent lane. It then accelerates and changes lanes to the right, eventually driving in front of the ego vehicle. & \includegraphics[width=6cm, height=3.7cm, trim= 0 -5mm 0 -7mm]{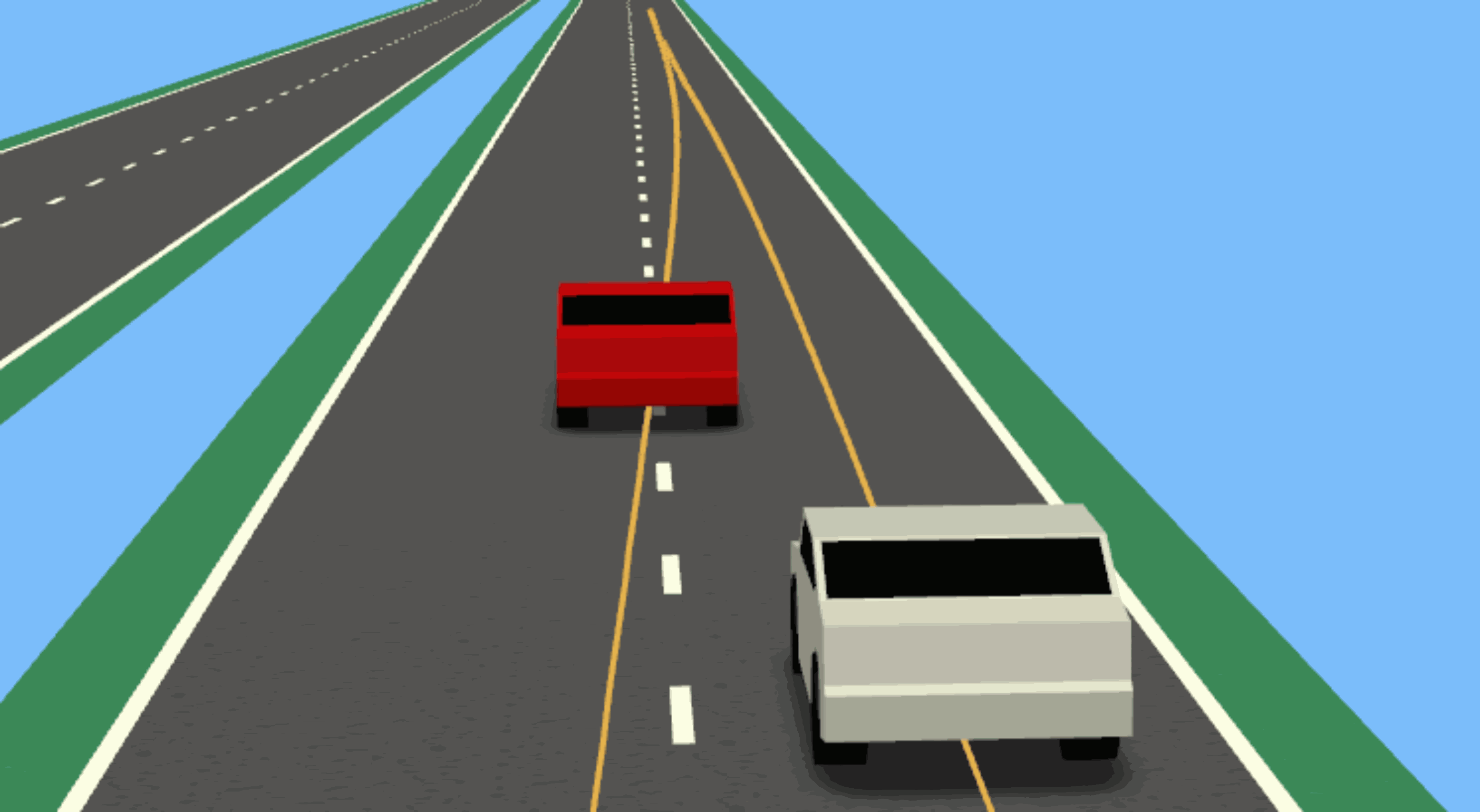} & \includegraphics[width=6cm, height=3.7cm, trim= 0 -6mm 0 -8mm]{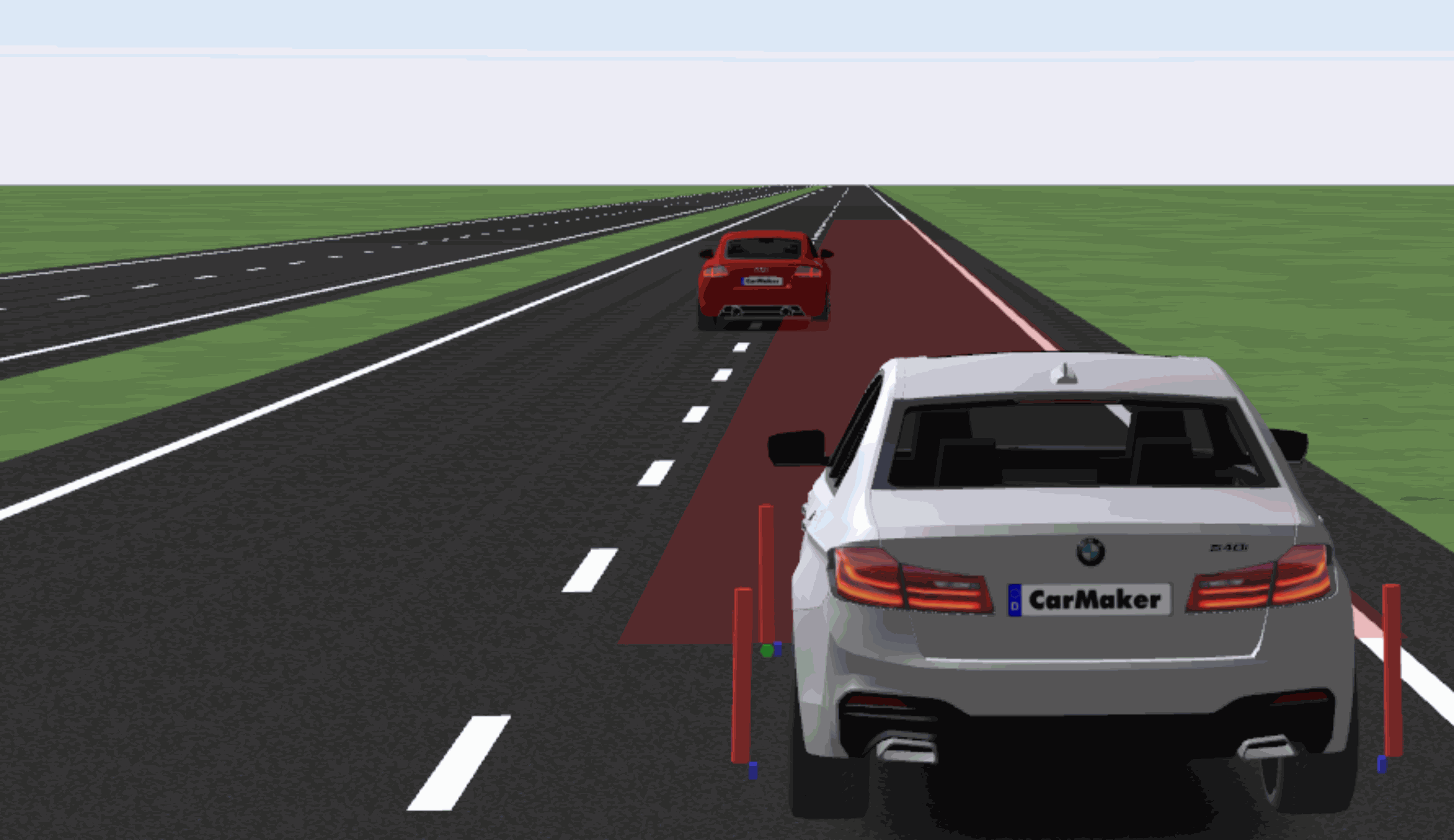} \\
\hline
\rotatebox[origin=c]{90}{Cut-out}&The ego vehicle follows the lane and maintains its velocity. Target vehicle \#1, initially driving in front of the ego vehicle in the same lane, accelerates and changes lanes to the right.& \includegraphics[width=6cm, height=3.7cm, trim= 0 -3mm 0 -5mm]{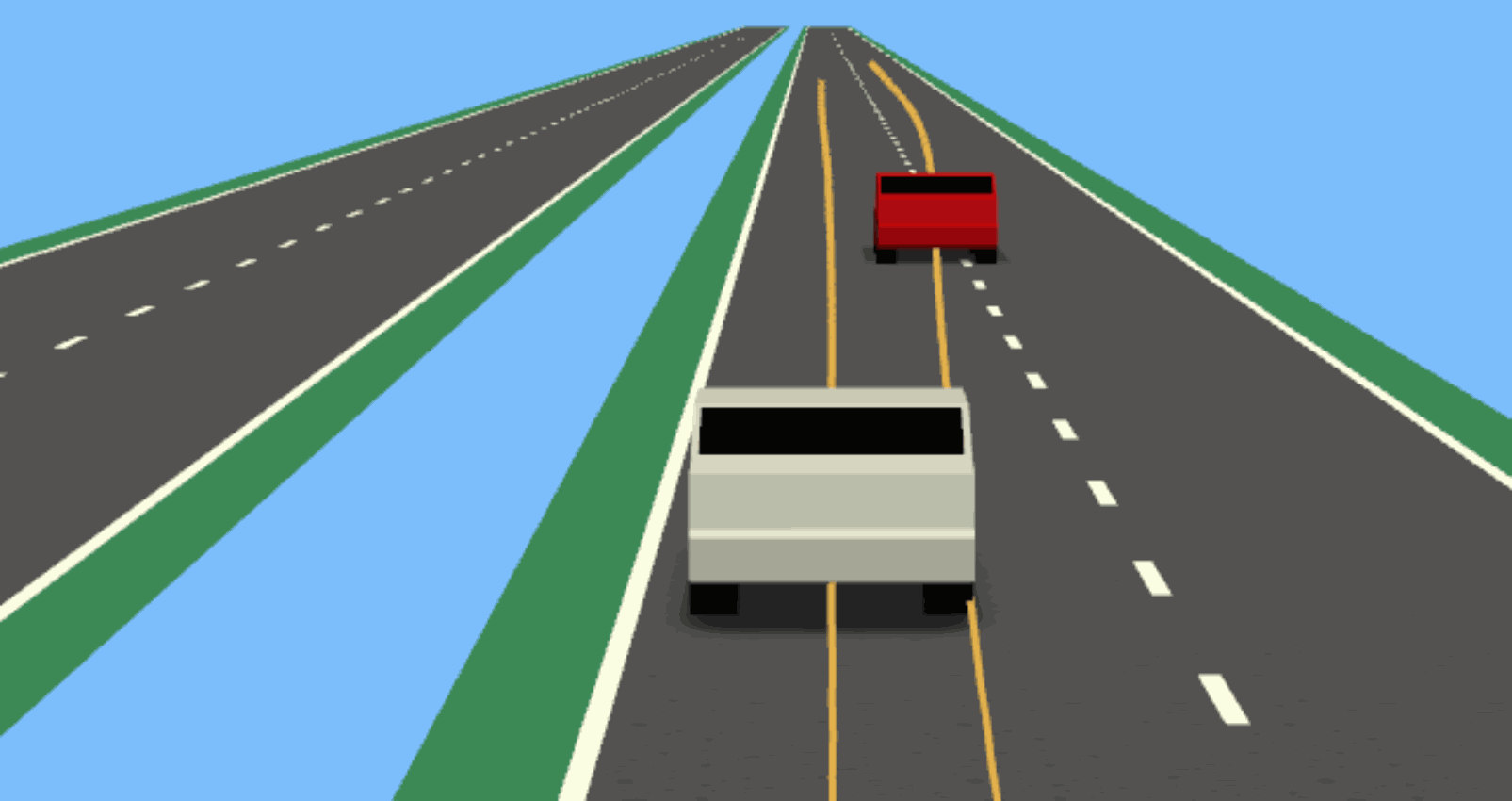} & \includegraphics[width=6cm, height=3.7cm, trim= 0 -6mm 0 -9mm]{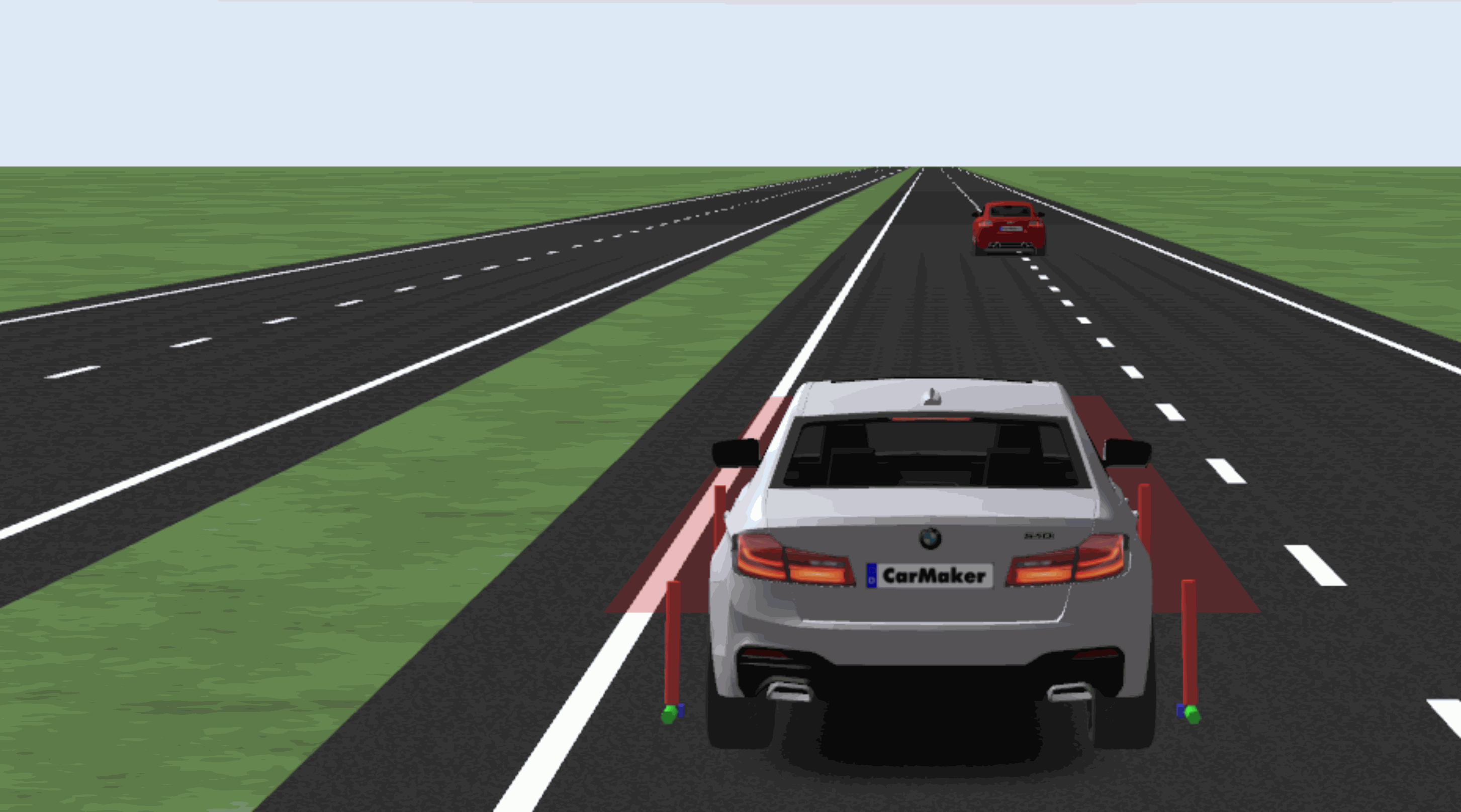}\\
\hline
\end{tabular}
\label{tab:case_study}
\end{center}
\end{table*}

The framework Chat2Scenario is validated through the extraction of three typical driving scenarios: following, cut-in, and cut-out scenarios. These scenarios are pivotal in testing ADS due to their frequency of occurrence and potential risk in daily driving. The following scenario tests the ADS capability to maintain safe following distances and respond to varying speeds of traffic. The cut-in and cut-out scenarios are critical for evaluating an ADS's lane-changing and overtaking strategies, as they involve the ego vehicle's reaction to other vehicles. These scenarios are successfully extracted from the dataset and reconstructed in Esmini and CarMaker, as detailed in~Tab.\ref{tab:case_study}. 

\subsection{Quantitative Evaluation}

In the highD dataset, ground truth labels for various scenarios are absent. To address this gap, semantic labels are manually generated through an analysis facilitated by a MATLAB-based visualization tool\footnote{\url{https://github.com/RobertKrajewski/highD-dataset}}. Given the intensive nature of the work, only one file - track \#36, which lasts about 27 minutes - is randomly selected from a total of 60 files for detailed analysis. In the labeling process, human driving experiences are used, but the vehicle's longitudinal activities are excluded due to the difficulty of manual identification from animated data. The quality of these labels is further verified by an independent reviewer. The effectiveness of the Chat2Scenario tool is assessed by comparing its outputs with the ground truth of track \#36, and the results are presented in~Tab.~\ref{tab:search_result}. 

\begin{table*}[htbp]
\caption{Quantitative evaluation of Chat2Scenario in track \#36}
\begin{center}
\begin{tabular}{|c|c|c|c|c|c|c|c|}
\hline
\textbf{Scenario Category} & \textbf{True Positive} & \textbf{False Positive} & \textbf{False Negative} & \textbf{Accuracy} & \textbf{Precision} &\textbf{Recall} & \textbf{F1 Score}\\
\hline
following & 2479 & 15 & 814 & 0.749 & 0.994 & 0.752 & 0.857\\
\hline
cut-in & 248 & 23 & 39 & 0.800 & 0.915 & 0.864 & 0.889\\
\hline
cut-out & 265 & 15 & 32 & 0.849 & 0.946 & 0.892 & 0.919\\
\hline
\end{tabular}
\label{tab:search_result}
\end{center}
\end{table*}

For the \enquote{following} scenario, a promising level of precision is demonstrated; however, this is mitigated by a significant rate of \enquote{false negative}, which can be attributed to the exclusion of scenarios that fall below the predefined duration threshold. In terms of the \enquote{cut-in} and \enquote{cut-out} scenarios, the values across all metrics indicate a robust competence in effectively identifying these scenarios. Overall, the capabilities of Chat2Scenario for scenario identification are substantiated by the quantitative metrics presented in the table.

\section{Conclusion and Future Work}

This paper has introduced Chat2Scenario, a publicly accessible web app that advances the extraction of concrete scenarios from naturalistic driving datasets. The platform interprets scenario descriptions in natural language and evaluates their criticality with precision, thereby streamlining the scenario generation process. The validity and the practicality of Chat2Scenario are substantiated through simulations in Esmini and CarMaker.

Comprehensive validation of the framework is to be primarily focused upon in future work to ensure its robustness and reliability. Additionally, the expansion of dataset diversity and the refinement of criticality metrics for customized evaluation are also planned.

\section*{Acknowledgment}

The authors would like to thank Dr. Hexuan Li for the discussion during the conception phase, Faris Orucevic for the technical support about CarMaker, and Zhengyu Wang for generating the ground truth labels.

\bibliographystyle{ieeetr}
\bibliography{Chat2Scenario}
\end{document}